项目编号　202310486133

# 武汉大学大学生创新创业训练计划项目结题报告

## 综合多模态空间数据的交通基础设施数字化关键技术研究

院（系）名　称：测绘遥感信息工程国家实验室

专　业　名　称　：测绘

学　生　姓　名　：田展源　朱添瑞　田泽睿

指　导　教　师　：董震　　　教授

二〇二三年二月

# INTERIM REPORT OF PLANNING PROJECT OF INNOVATION AND ENTREPRENEURSHIP TRAINING OF UNDERGRADUATE OF WUHAN UNIVERSITY

## Research on Key Technologies of Infrastructure Digitalization based on Multimodal Spatial Data

| | | |
|---|---|---|
| College | : | Wuhan University |
| Subject | : | Geomatics |
| Name | : | Zhanyuan Tian, Tianrui Zhu, Zerui Tian |
| Director | : | Dr.Zhen Dong |

March 2013

# 郑 重 声 明

  本项目组呈交的中期报告，是在导师的指导下，独立进行研究工作所取得的成果，所有数据、图片资料真实可靠。尽我们所知，除文中已经注明引用的内容外，本报告的研究成果不包含他人享有著作权的内容。对本报告所涉及的研究工作做出贡献的其他个人和集体，均已在文中以明确的方式标明。本报告的知识产权归属于培养单位。

  项目组签名：_____________　　　　日期：_____________

# 摘 要

自 2010 年 NASA 提出数字孪生概念以来，许多产业均提出数字化发展的动态目标，交通产业也在其中。随着越来越多的公司在这块处女地上布局，数字孪生交通产业迅猛发展，并逐渐形成了完备的科学研究体系。但在大体成熟的框架下，仍有许多漏洞问题亟待解决。

在用点云信息构建路网的过程中，我们总结了激光扫描仪采集的点云的几大特点，并分析了潜在构网问题，如错判地物点为地面点、网格空洞等。在此基础上，我们查阅相关文献，提出了具有针对性的解决方案，如仿照影像金字塔搭建点云金字塔、扩展虚拟格网等，应用 CSF 进行地面点云提取，并利用 PTD（Progressive-densification-basedfilters）算法构建路网模型。。

针对路牌检测的问题，我们通过使用边缘检测增强信息密度、通过去除低强度点提高数据质量以及利用 PaddleOCR 和 Densenet 实现道路文本识别的 90% 准确率，优化地面点云中的遥感数据。

而在数字孪生交通的实时化方面，我们设计 P2PRN 网络——利用 MPR-GAN 的骨干进行 2D 特征生成，利用 SuperGlue 进行 2D 特征匹配，根据匹配优化点渲染视角，经过多次迭代完成多模态匹配任务，并成功计算道路摄像机位置 ±10° 和 ±15m 精度。

**关键词**：数字道路孪生；PTD 加密算法；目标检测；P2PRN；2D-3D 多模态数据匹配

# ABSTRACT

Since NASA put forward the concept of the digital twin in 2010, many industries have put forward the dynamic goal of digital development, and the transportation industry is also among them. With more and more companies laying out on this virgin land, the digital twin transportation industry has grown rapidly and gradually formed a complete scientific research system. However, under the largely mature framework, there are still many loophole problems that need to be solved.

In the process of constructing a road network with point cloud information, we summarize several major features of the point cloud collected by laser scanners, and analyze the potential problems of constructing the network, such as misjudging the feature points as ground points and grid voids. On this basis, we reviewed relevant literature and proposed targeted solutions, such as building a point cloud pyramid modeled after the image pyramid, expanding the virtual grid, etc., applying CSF for ground-point cloud extraction, and constructing a road network model using the PTD (progressive density-based filter) algorithm.

For the problem of road sign detection, we optimize the remote sensing data in the ground point cloud by enhancing the information density using edge detection, improving the data quality by removing the low intensity points, and achieving 90% accuracy of road text recognition using PaddleOCR and Densenet.

As for the real-time digital twin traffic, we design the P2PRN network using the backbone of MPR-GAN for 2D feature generation and SuperGlue for 2D feature matching, rendering the viewpoints according to the matching optimization points, completing the multimodal matching task after several iterations, and successfully calculating the road camera position ±10° and ±15m accuracy.

# 目　录

# 第 1 章　绪论

## 1.1　研究背景及意义

数字孪生交通旨在在网络虚拟空间中实现对现实世界的人、车、路等全要素的实时数字化和可视化，既包含静态道路基础设施三维重建（如，位置信息、语义信息和拓扑信息），又包含动态实时交通信息（如，人、车、非机动车的状态、速度、方向、位置等）精准映射，需要综合利用多源、多平台传感器进行数据采集。2019年9月19日，中共中央、国务院印发《交通强国建设纲要》，明确了前沿技术与交通运输全面融合的战略目标，提出了对交通自动化、数字化、智慧化在新时代下的要求，为数字孪生交通的蓬勃发展提供了政策指导基础。

同时，我国车辆数量的激增使城市对拥堵评判、指挥调度、智慧交通管制等方面产生了巨大需求，为数字孪生交通产业提供了现实发展温床。许多城市早已着手数字交通的建立，如上海嘉定计划在未来新建50个"全息路口"，路口两侧的综合杆上除了标配的电子监控设备外，还搭载了多套电子设备——如雷达、信号发射器等，承担着数据采集、接收和信息发布等功能，为车辆的行驶提供超视距感知。在这条路上，流量、车速、轨迹等各类通行数据都被摄像头捕捉，助力交通管理部门实现通行环境提前预警、交通组织优化、信号灯自动切换等智慧交通管理，让城市交通更加便捷且安全。放眼全世界，自2022年来，苹果、谷歌等巨头斥资推进交通数字化研发，英国交通总局在计划在未来五年内完成道路数字化的演变[27]，越来越多的公司都在这块处女地上布局，数字孪生交通也在全球科技舞台上大放异彩。

## 1.2　研究方法

将道路和交通流信息数据化需要以下几个步骤：

首先应利用采集到的道路点云和影像信息构建各个道路的路网模型；然后根据图像中检测到的路牌文字信息，在上述路网模型中添加指示以及周遭环境信息；最后，将摄像头检测到的实时车辆流信息镶嵌进路网模型中，完成实时数字道路的构建。

在第一个部分中，如何利用采集到的道路点云信息构建路网模型是一大挑战。我们团队采用了基于地表的点云滤波进行预处理，然后对剩下点云进行三角剖分。

为了尽可能地拟合不规则的地面形状、同时保证构网时能保留约束项，我们采用 PTD（基于插值法的不规则三角网渐进加密）进行构网，并针对构网侵蚀地面、网格产生空洞等问题提出了改进策略。

在第二个部分中，难点在于路牌信息的检测。尽管如今目标检测算法已经十分成熟，在复杂的道路环境下检测出路牌仍是一个挑战。因此，我们团队采用了一步式检测的 YOLOv5 算法检测路标位置，在多张像片同时出现的路牌，由于已知拍摄像片位姿，即可用前方交会的思想确定路牌在三维空间中的位置。

最后，在第三个部分中，道路的实时数字化需要借助摄像头信息——通过摄像头获取的二维图像检测出车辆，然后将车辆位置信息投影到构建完成的路网模型中，并由已成熟的跟踪算法（Tracking）技术统计总车流量。以上过程的关键在于如何在未知摄像头位姿（Pose）的前提下，将摄像头影像上的车辆坐标换算成真实世界坐标。我们团队采用多模态数据融合的思路，将二维的影像和三维的点云进行匹配（2D-3D Match）：由于前期获取的点云均已知其世界坐标，若能将摄像头影像中部分特征和点云信息匹配，即可得到影像坐标和世界坐标转换关系。最后利用单应变化公式将车辆位置投影（warp）到路网模型上，即可实现基于路网模型交通流检测。其中单应矩阵计算公式如下：

$$H_2(d) = K_2 \cdot R_2 \cdot \left(I - \frac{(t_1 - t_2) \cdot n_1^T}{d}\right) \cdot R_1^T \cdot K_1^T$$

（1.1）

## 1.3 主要贡献和创新点

在构建路网模型时，我们结合使用了数据金字塔和虚拟节点，尽可能地避免了三角网建造过程中的空洞和错判地表点。

在对路网模型实时化的过程中，我们把摄像头影像信息和路网模型的融合视作多模态数据融合，并突破性地在图像（二维）层面实现了二维-三维特征匹配。同时，在后续纠正摄像头位姿的过程中，我们巧妙利用了对抗神经网络（GAN），用采集到的道路点云数据生成出了以假乱真的另视角图片。而在后续匹配过程中，我们又连接了点云渲染和特征匹配两大领域，并在训练中分两阶段训练，最后再迭代纠正位姿，提高了特征匹配的精度。

最后，通过我们的工作，数字孪生交通能拥有更加完整和精准的道路构网，同

- 2 -

时，数字交通中的实时路况信息也能更精确地融合进上述静态构网中，为后续数字交通在实际中的应用打下坚实的基础。

# 第二章 路网模型

## 2.1 路网模型的建立原理

路网模型是数字孪生建模与分析的基础，有精确精细的路网模型才能使后续的工作平稳准确的展开，其中生成路网模型的原始数据来源广泛，包括众源数据、OSM 数据等，而机载/车载激光雷达生成的点云数据可以直接记录从地物或地表返回的密集、离散、细节丰富、精确的三维点云，那么如何使用这些不规则点云生成路网模型是我们要解决的问题。

## 2.2 相关工作

道路三维基底包括精准的三维几何信息和精细的拓扑关系信息。在精准三维几何模型重建方面，根据模型表达形式分为体素模型、三角网模型及隐式函数模型等。SurfaceNet 通过预测场景中体素位于物体表面的置信度来重建场景表面体素模型（Ji et al., 2017）。RayNet 通过提取视点无关的特征和显式编码几何约束重建场景体素模型（Paschalidou et al., 2018[7]）。基于三角网的模型重建方法通过节点、边和面片构建场景表面模型，一些方法将该任务抽象为从影像(Wang et al., 2018[4]) 或点云(Ladicky et al., 2017[12]) 中生成三角网。三角网模型依据道路平面、纵横断面的设计结果，模型能够准确计算横断面各个角点的 3D 坐标，将相邻横断面对应角点连接形成三角网。孔斯曲面通过连接大量曲面片组成具有复杂特征的曲面，所有用于曲面构造的曲面片均由四条边界确定，可以实现面和片之间的光滑连接。其中关于三角网构建的理论包括基于动态算法构建 CDT （三角网）理论， 赋予数据点、三角形面相应的属性，快速完成拼合交线点建网，以此作为约束条件嵌入三角网内，赋予拼合交线区域内的三角形设计面属性，赋予区域外三角形地表面属性。最后通过将建立的模型和三维点云导入同一场景中，借助点云可进行模型与实际地物、地面信息位置校对，完善模型的建立。与前两者的显式模型不同，隐式模型没有将三维模型显式地重建于场景中，而是通过深度自编码器自主学习高维特征，再将场景信息，包括几何、语义、纹理等，编码至网络模型中隐式表达(Sitzmann et al., 2019[2]; Tobin et al., 2019[3])。

在精细的道路拓扑关系重建方面，常见的方法是以图模型实现中心线级路网拓扑重建、车道级路网拓扑重建、车道连通性与虚拟车道重建等任务。对于中心线级路网拓扑重建任务，Mattyus et al. (2015)[5]融合 Open Street Maps 中的道路、人行道及停车场等信息，通过图模型生成路网；Ventura et al. (2018)[6]通过卷积神经网络预测局部区域中与已道路中心点连通的边界道路点，得到全局路网的拓扑结构。针对车道级路网的提取，DAGMapper 通过将车道线的局部几何和拓扑信息编码至图节点中，然后通过贪婪算法推导大范围车道拓扑，最终实现高速公路上的多车道模型重建 (Homayounfar et al., 2019)[8]。为进一步提高复杂路口处的路径规划和导航精度，Ye et al. (2020)[9]重建了路口处的虚拟车道，先根据路面标志插值出车道中心线，再根据路口处的停止标线区分车道中心线的起始和终止节点，以构建路口处的车道连通关系与虚拟车道。在现有的软件方面，英国开发的 MXROAD 是道路设计系统中较具代表性的一款，可以满足道路复杂设计问题的需要，如交叉道口设计、环岛、桥隧以及路面标线等。

在使用点云创建路网模型时国内外创建出多种处理方法用于滤波，典型的有 slope-based、block-minimum、surface-based、clustering/segmentation algorithm 等。

Surface-based 的核心步骤是创建一个最接近裸露地表的表面，其中最经典使用最广泛的为三角剖分算法，即通过对点云数据进行自动化预处理，地面滤波，保留地面点，剩余的地面点通过构建不规则三角网（TIN）模型进行栅格化，可得到高精度的地面数字高程模型。德洛内(Delaunay)三角网对剖分结果具有良好的控制力，其优点是结构良好，数据结构简单，数据冗余度小，存储效率高，与不规则的地面特征和谐一致，可以表示线性特征和迭加任意形状的区域边界，易于更新，可适应各种分布密度的数据等。因此建立路网的两大过程为点云滤波和三角网建立。

## 2.3 我们的工作

根据激光扫描仪的特性，其扫描结果会有不完整的点云，获取信息质量差，此时需要对点云进行处理（处理方法包括去噪、简化、分割、合并等）并对处理后的结果进行约束建立带有约束附加项的三角网。结合以上需求，我们使用改进后的不规则三角网渐进加密算法建立道路模型三角网。

PTD, Progressive-densification-based filters, 即基于插值法的不规则三

角网渐进加密算法，对各种地形均有良好的表现。其主要分为三个步骤，去除孤立点、选择种子点并构建初始 TIN 模型、迭代加密 TIN 模型。PTD 性能在不连续地形的地面提取的总误差综合性能是最优的，但是在复杂地形处 PTD 不仅容易侵蚀地面，还容易对低位地物点误判为地面点，这导致网格中含有空洞或不合理三角形。针对以上问题，我们进行了如下改进：

1. 使用数据金字塔，以不同尺寸的虚拟格网组织点云，与迭代相结合。具体策略是三维空间中将点云投影在二维平面后把 XOY 平面分割为正方形网格，选取格网中的最低点作为种子点，每一次迭代时网格尺寸变为原本的二分之一，重新构建的格网获得更高的分辨率，形成点云金字塔，知道满足停止条件

2. 扩展虚拟格网，插值虚拟种子点，有效补充边界区域。我们在点云测区的四个角添加模拟角点辅助构建 TIN，或者对参与构网的种子的点向外扩展延伸。

3. 采取附加项两边夹角约束，法向量约束等条件，如使用向量乘积结果控制两边夹角的约束，使其不要成为钝角，即

$$\begin{cases} \overrightarrow{P_2P_3} \cdot \overrightarrow{P_2P_4} > 0 \\ \overrightarrow{P_2P_3} \cdot \overrightarrow{P_2P_4} = 0 \\ \overrightarrow{P_2P_3} \cdot \overrightarrow{P_2P_4} < 0 \end{cases}$$

（2.1）

并且为防止重建表面几何特征变化过于剧烈，对网格法向量的夹角进行计算，使其小于 90 度。

算法具体流程为对初始激光点云进行预处理，们用 KD-Tree 组织点云数据，遍历点云，搜索每个点在指定范围内的临近点数，当点数小于给定阈值，则判定为异常值，剔除异常值；构建数据金字塔，初次使用大于最大建筑物尺寸的格网，依次迭代缩小格网尺寸，形成点云金字塔；使用 PTD 滤波，根据距离阈值和角度阈值，判断地面点，这个过程中使用改进的虚拟格网方法后对附加项进行约束连接所有的地面点形成三角网。

本次实验使用两个数据集，分别用于滤波和构成三角网。考虑到激光采集点云时会包括低矮灌木，数据集 1 使用斜坡上的道路数据含有低矮植被噪声，

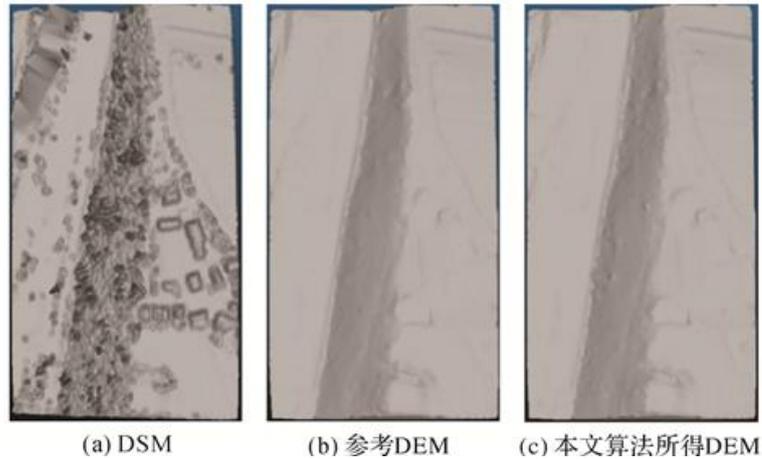

图 **2.1**　所得 DEM 示意

由比对结果看来效果理想。

第二段实验将会采用实际激光点云测试数据，但目前还在寻找合适数据集，为下一个阶段开展的工作。

# 第三章　路标检测

## 3.1　路标检测原理

路标信息以及交通标识牌的信息是交通系统中非常重要的部分，随着近几年深度学习的发展，通过目标检测算法就能很快地提取出相应的信息。基于深度学习的目标检测方法主要是以下两个方向：基于候选区域的 R-CNN（Regions with CNN features）系列的目标检测算法，其基本思想是首先提取目标的兴趣区域，然后根据图像中的一些特征信息使用分类器对目标进行分类；基于端到端的 YOLO（You Only Look Once）目标检测算法。首先将输入图像划分为若干个个网格，在网格内需要预测目标的候选框和类别信息这两个信息，YOLO 算法[28-30]可以一次预测所有区域包含的目标候选框、类别概率以及目标置信度，直接输出目标的检测与分类结果。路标信息提取整体任务分成检测和识别两个阶段，在检测阶段识别出交通标识牌的类别，而路标的语义信息则通过一个单独的文本识别模型提取。由于交通标识牌及商家店名等文字信息在实际应用中存在排布形式与方向多样、拥有多种语言混合的问题，普通的识别模型显然难以解决。因此，我们尝试引入了最早于机器翻译领域提出和应用的 Attention 机制，在卷积过程中对目标字符以及其附近像素

赋予更大的权重，使得识别模块的"注意力"更加集中对应到目标字符，从而实现较长输入序列的合理向量表示。

## 3.2 相关工作

在以前的大多数工作中，传统的基于机器学习的方法，如 HOG 特征和支持向量机，只是为了检测特定类别的交通标志而设计的。考虑到交通标志固有的颜色和形状信息，还会使用圆形检测算法和基于 RGB 的颜色阈值技术来检测圆形交通标志。然而，无论是直观的颜色特征还是直观的形状特征在真实复杂的路面环境下都不可靠。因此本文主要讨论用卷积神经网络（CNN）解决检测任务。

YOLOv5（You Only Look Once）是由 UltralyticsLLC 公司于 2020 年 5 月份提出。YOLO 算法用一个神将网络直接把输入图像转换为包含预测物体的信息的张量。YOLO 算法相较于其他目标检测算法更加快速，非常适合在交通场景下使用。YOLOv5 算法和上一代相比能够完成多目标小目标检测任务，这也是适合交通路标检测任务的一个重要原因。

自然场景文本检测识别（Scene Text Recognition, STR）是机器视觉中的一个重要领域。它在识别商品、照片、视频、还有我们这次需要识别的路标中有着重要作用。相比于已经比较成熟的文档文本识别技术（OCR），STR 显然更具挑战性。自然场景下有着复杂的光照、倾斜、遮挡、以及文字排版等多个因素干扰。因此传统的 OCR 并不能满足我们的需要。

在将校正方法引入 OCR 之后，R2AM 算法首次将 Attention 机制引入文本识别领域。该模型首先将输入图像通过递归卷积层提取编码后的图像特征，然后利用学习到的统计信息通过 RNN 解码输出字符。在此之后，在 Attention 领域进行探索和创新的算法越发多样和高效，对误差传播、训练成本等方面进行了很大的优化。但很多基于 Attention 的方法在处理像素较低或文本排版复杂的图像时，无法将这种特殊图像中字符的注意中心准确地集中到对应的目标区域的中心，因而导致算法表现不尽如人意。

## 3.3 我们的工作

本文的实验数据集选用清华腾讯联合制作的交通标志数据集 TT100K，其中有 10000 张图片和 30000 个交通标志，大小均为 2048×2048。部分样本如图 3.1 所示。数据集中小目标是最常见的，在 TT100K 上进行检测非常具有挑战性。考虑

到某些类别的交通标志个数太少不利于网络学习，对数据集进行划分，多于 100 张的分类被保留，再按照 7：2：1 的比例划分成训练集、验证集和测试集。最后训练集 6793 张，验证集 1949 张，测试集 996 张，涵盖了 45 种常见交通标识。

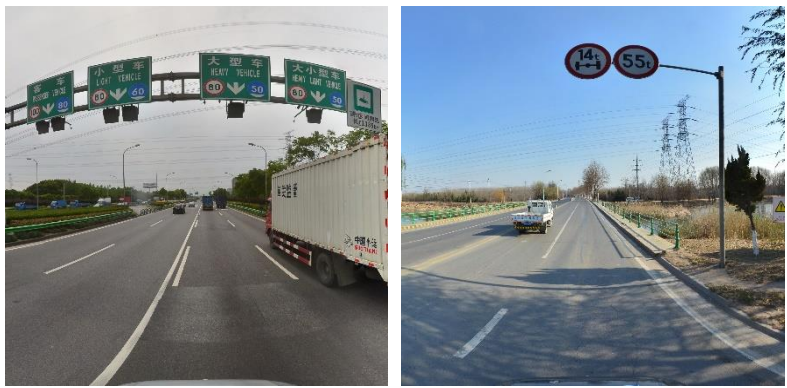

图**3.1**　部分数据集

设置好输入输出（IO）后，将数据集导入 YOLOv5 模型开始训练，其中 batch_size 选取为 64，Epoch 设置为 1000 轮。

实验结果大部分识别准确。但是较远位置较小的目标难以识别，并且原始图像比 resize 后的图像更易识别，因此对于小目标的检测做进一步的数据增强是有必要的。

在阅读大量文献之后，发现在深层特征提取中小目标的特征融合后被削弱甚至消失，因此在特征融合的阶段应增加权重抑制深层特征，加强浅层的小目标特征，以帮助实现小目标的检测。由此我们采用了 MASTER 中的核心模块，即 Multi-Aspect 全局上下文建模方法[31]，纠正 Attention 注意中心漂移问题，大大提高了文本识别精度。

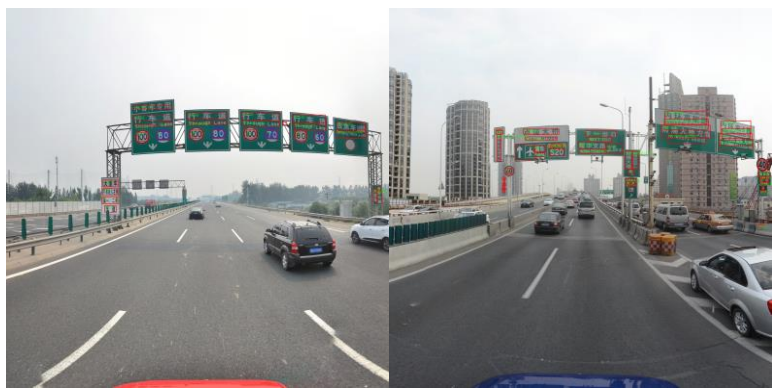

图 **3.2**　部分结果图

- 8 -

# 第四章 影像-点云匹配

## 4.1 影像-点云匹配原理

之前提到，数字交通实时化的关键在于摄像头实时影像和静态路网模型的融合，而这需要高精度的影像-点云匹配技术。过去二十年中，基于图像的二维-二维特征匹配技术高度发展，为了获取二维特征，工业界开发了诸如 SIFT[1]、SURF 等高效特征提取算法。而与传统的二维-二维特征匹配相比，二维-三维特征匹配需要跨模态进行匹配，因此需要考虑如何将多模态数据融合到同一领域进行匹配。

目前主流的融合方法有两种：一是将点云和影像分别通过两个特征提取网络，得到同维度的特征信息，在特征空间进行匹配；二是对多视角（Multi-views）图像进行稀疏点云重建，再利用三维特征提取算子分别提取重建点云和原点云特征，在三维空间上进行匹配。

## 4.2 相关工作

Feng et al.（2019）[10]首次提出使用深度学习方法学习特征描述，实现在二维图像和三维点云中直接匹配关键点，该方法首先使用 SIFT（Lowe, 2004）[1]算法和 ISS（Zhong, 2009）[13]算法分别在图像和点云中提取关键点，然后使用深度学习方法分别学习对应的特征描述符，最后在特征空间中获取 2D 和 3D 关键点之间的对应关系，从而建立来自查询图像和三维参照点云的 2D 到 3D 对应。Uzair et al.(2019)通过网络训练描述子匹配，在点云和图像中预先提取关键点，通过描述子匹配算法寻找同名点对。Pham et al.（2019）[11]使用局部双重自动编码神经网络同时将图像和点云映射到共享的隐藏空间，该方法相对于单独在图像和点云中分别学习特征描述，描述子更具区分性，有利于 2D-3D 匹配。然而该方法只能针对图像与点云的局部区域进行匹配，难以适应大范围场景的 2D-3D 匹配任务。

## 4.3 我们的工作——P2PRN

从前人的工作中不难发现，二维-三维特征匹配的两种策略--"在特征空间进行匹配"和"在三维空间上进行匹配"均存在难以解决的问题：前者的工序明显缺乏可解释性，且整个匹配过程重度依赖特征提取网络，这会导致模型缺乏泛化性；而后者工序繁琐，且缺乏准确、鲁棒的三维特征提取方法，准确性较低。

为了突破当下多模态数据匹配困境，我们团队创新性地提出了在二维图像层

面进行匹配的算法——基于点渲染的像素-点云匹配策略（Pixel and Point Matching Strategy with Point Rendering Neural Network，下面简称 P2PRN），其基本流程图如下。

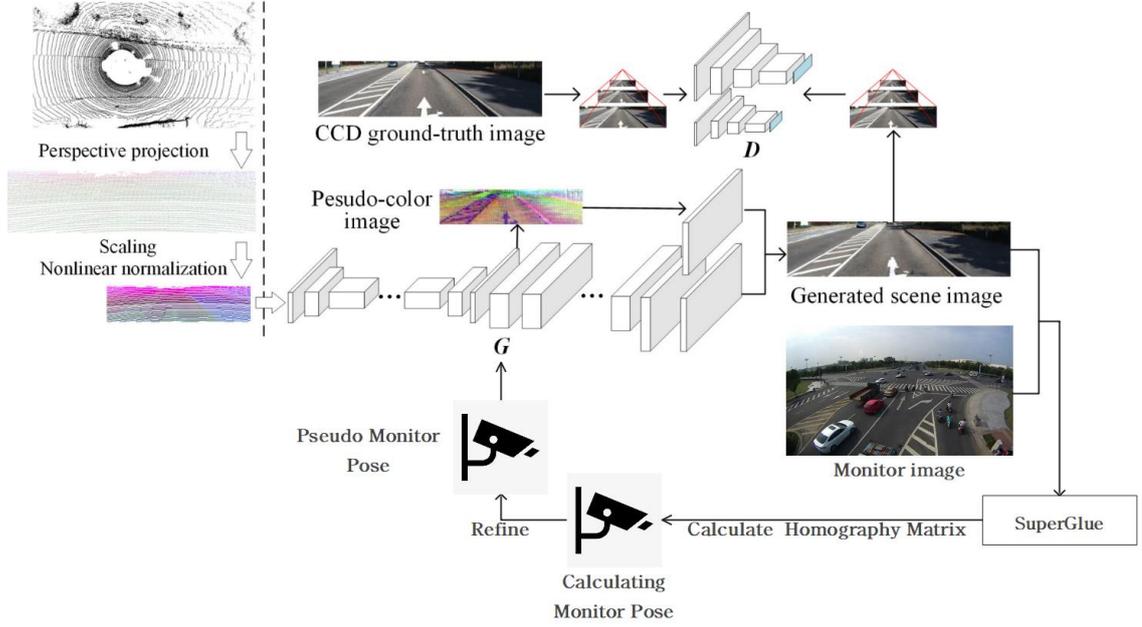

图**4.1** P2PRN算法流程

其中，点渲染网络的原型为 Qingyang Xu 等人研发的 MPR-GAN 网络[22]：将采集到的点云数据输入 GAN 网络的生成模型（G）中，生成场景图片后，将其与真实场景图片一起输入判别模型（D）中，最后采用论文中的杂损函数（Morbid Loss Function）进行训练。

在此网络的基础上，我们引入道路摄像头的估计位姿(Pseudo Monitor Pose)，利用生成模型输出虚拟视角下的路网点云模型像片（Generated Scene Image）。然后，对虚拟影像与真实道路摄像头影像进行特征点匹配（此处采用 SuperGlue 作为特征匹配算法[23]），再根据式（1.1）计算真实摄像头影像和生成摄像头影像间的单应变换矩阵，最后据此修正道路摄像头的估计位姿。

$$H_2(d) = K_2 \cdot R_2 \cdot \left(I - \frac{(t_1 - t_2) \cdot n_1^T}{d}\right) \cdot R_1^T \cdot K_1^T$$

（4.1）

式中 $K_2$、$K_1$ 分别代表道路相机和车载相机的内参，$R_2$、$R_1$、$t_1$、$t_2$ 分别为道路相机和车载相机的旋转矩阵和位移矢量。

迭代上述过程若干次，直到道路摄像头估计位姿的修正值小于临界值时停止，

即可得到较为准确的摄像头位姿。据此,我们即实现了二维-三维特征跨模态匹配。再用式(1.1)代表的单应变换,即可将实时的路况信息投影到静态路网模型中。

## 4.4 模型训练和实验

在具体的训练细节上,我们拟采用分段式训练(Two-stage Training):首先将道路点云和影像喂给 MPR-GAN 网络,损失采用原论文中的杂损函数(Morbid Loss Function),训练出高精度的生成模型网络。然后再使用不同视角的道路影像(已知拍摄位姿)训练 SuperGlue 网络中的注意力神经网络和匹配神经网络(以下简称 SuperGlue 网络),将不同视角间真实的单应变换矩阵和 SuperGlue 估计出的单应变化矩阵做二范数损失(L2 Loss)。最后用训练完成的 SuperGlue 网络对摄像头视角的生成相片和摄像头真实影像进行特征匹配,在此期间固定 MPR-GAN 网络和 SuperGlue 网络参数不变,根据先验摄像头位姿和计算出的摄像头位姿之差,迭代至精度达标。

在实验中,我们将 P2PRN 网络分成两个部分进行测试。

第一个需要测试的部分是点渲染的精度。针对 P2PRN 前端点渲染网络,我们复现了 MPR-GAN 网络的结果,在 KITTI 数据集上完成了测试(部分结果摘录自文献[22]),我们将真实背景图片作为真值,计算渲染得到的图片的 PSNR(峰值信噪比)和 SSIM(结构相似性)数据,结果如下表所示。

表 4.1 P2PRN 点渲染网络结果评定

| MODELS | PSNR | SSIM |
| --- | --- | --- |
| NPCR-MP-Z2P | 11.349 | 0.348 |
|  | 14.086 | 0.378 |
| LF-FCN | 14.723 | 0.405 |
| Pix2PixHD | 15.574 | 0.367 |
| P2PRN | 16.906 | 0.424 |

不难看出,就 PSNR 指标而言,前端点渲染的结果优于 NPCP-MP 等一众点渲染网络的结果,而就 SSIM 指标而言,P2PRN 也领先大部分网络。这足以证实我们点渲染得到的生成图片的可靠性。

同时,我们先简单地选取了车辆行驶视角进行测试,尝试通过匹配渲染图像和真实图像间的特征点计算出像片拍摄时相机的相对位姿。我们选取车辆行驶第十时间刻拍摄图片作为摄像头视角,选取车辆行驶第五时间刻拍摄图片作为初始迭

代视角，第一轮迭代的可视化结果如图 4.2 图 4.3 所示：

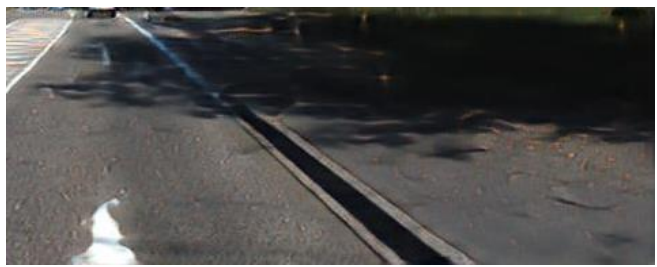

图 **4.2**　　第一轮点渲染结果

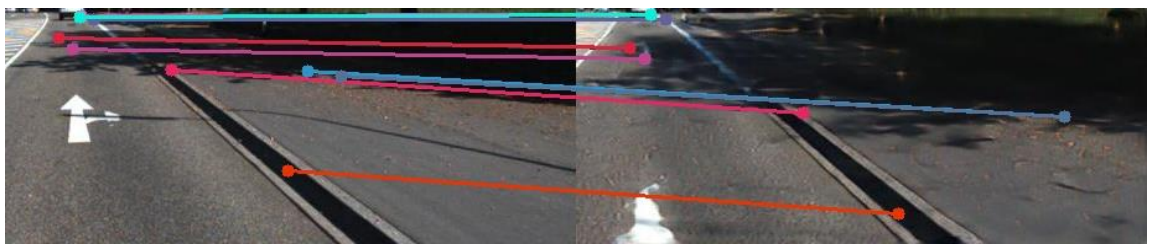

图**4.3** 第一轮迭代（左图为真实图像（摄像位置），右图为渲染图像）

可以看到除去少量匹配错误（如橙线）外，绝大多数匹配均是合理的，但我们也可以看到，由于大多数匹配均依赖于树荫等时效性特征，这要求我们需要选取与道路相片拍摄时间刻相同的摄像头时间刻进行匹配。

计算出二者相对位姿后，我们将相机移动至新的位置，并渲染出更新后视角的图像，并与真实图像进行特征匹配，第二轮迭代的可视化结果如图 4.4 图 4.5 所示

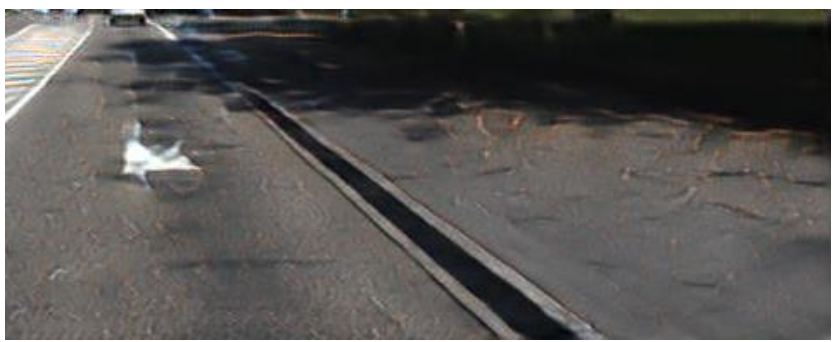

图 **4.4**　　第二轮点渲染结果

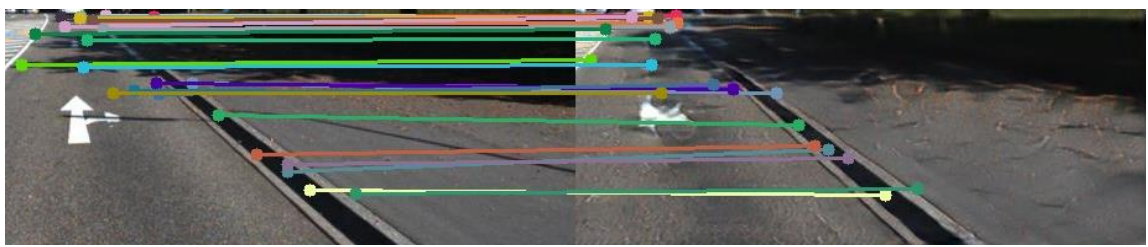

图**4.5** 第二轮迭代（左图为真实图像（摄像位置），右图为渲染图像）

可以看到相比第一轮迭代，渲染图像明显更加像真实图像了，证明我们的位姿迭代是有效的。我们设定当渲染图像迭代前后变化率小于 5%为迭代停止条件，当这样多轮迭代后，我们最终可以计算出左图拍摄时的位姿。

第二个需要测试的部分是迭代得到的摄像头位姿精度，我们根据匹配结果不断迭代渲染图像使用的位姿，最终计算出了左图拍摄时的相机位姿，并将其与真实位姿进行比较，结果基本一致，确定了我们匹配策略的可行性。

# 第五章　路面文字识别

## 5.1　路面文字识别原理

随着智能交通系统及无人驾驶技术的快速发展，其必要组成部分——高精地图（HD Map, High Definition Map），被认为是未来出行的关键一环，是交通资源全时空实时感知的载体和交通工具全过程运行管控的依据。其需要具备几何精度高、新鲜度高等特点。而在传统交通领域，为了给驾驶员方便快捷的提供准确的导航信息和交通指示，许多国家和地区在道路上使用文字标识来指示方向、距离和目的地等信息。因此，路面文字是高精地图中不可或缺的关键信息，对于路面文字的采集和提取由此成为了高精地图制作中的一个重要问题。

近年来，搭载多种传感器（包括激光扫描仪、GNSS（Global Navigation Satellite System）、IMU（Inertial Measurement Units）等）的移动测量系统，可同时获取道路及路侧地物表面的高精度三维几何信息和属性信息（如反射强度、回波次数等），为道路场景理解、高精度三维重建、空间分析等提供了重要的数据源。我们结合点云信息及字符识别模型，给出了我们对这个问题的解决方案。

## 5.2　相关工作

目前对于路面文字的提取和分析仍然存在一些挑战。传统上，路面文字的识别主要依赖于人工的观察和解读。这种方法存在人力成本高、效率低和容易出错的问题。而基于车载相机采集的视频和序列影像易受到光照和天气的影响，且缺乏准确的高程信息，难以满足三维高精地图的要求。

因此，需要提供一种新的技术和方法，能够高效、准确地提取路面文字，并应对各种复杂情况。因此，本小组创新性的提出了从点云文件中经由传统图像处理算法得到有效图片再进行识别的工作流程。但受激光扫描机制的影响，点云反射强度

受地物表面材质、扫描距离及入射角影响，给精确提取路面文字带来挑战。

## 5.3 我们的工作

### 5.3.1 点云文件处理

点云文件由激光扫描得到，包含了真实的点坐标以及反射率等信息，在空间上具有离散性、无序性的特点。而对路面文字进行识别只需要其中的二维信息，因此将点云文件进行地面滤波后就能转换为二维影像信息，再通过后级网络进行识别。

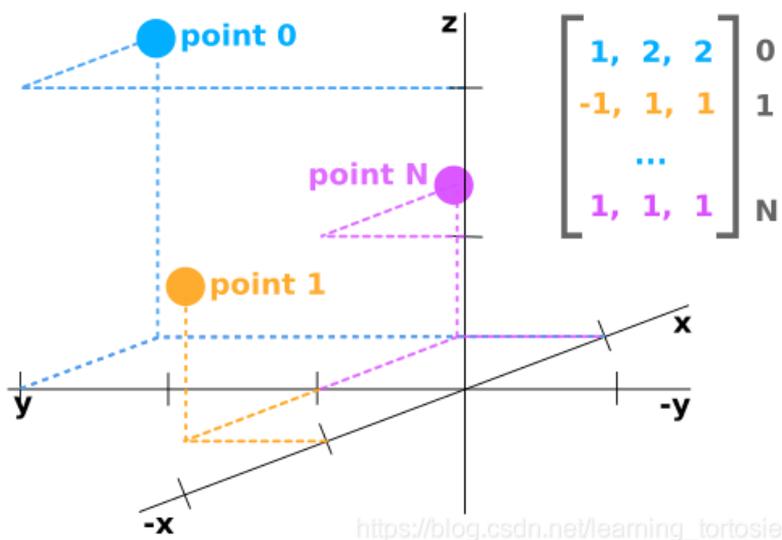

通常来说遥感影像的分辨率极高，而信息密度低，过高的分辨率与尺寸会造成网络溢出。因此加入滑窗边缘检测算法划分小图片并将无信息区域删除。

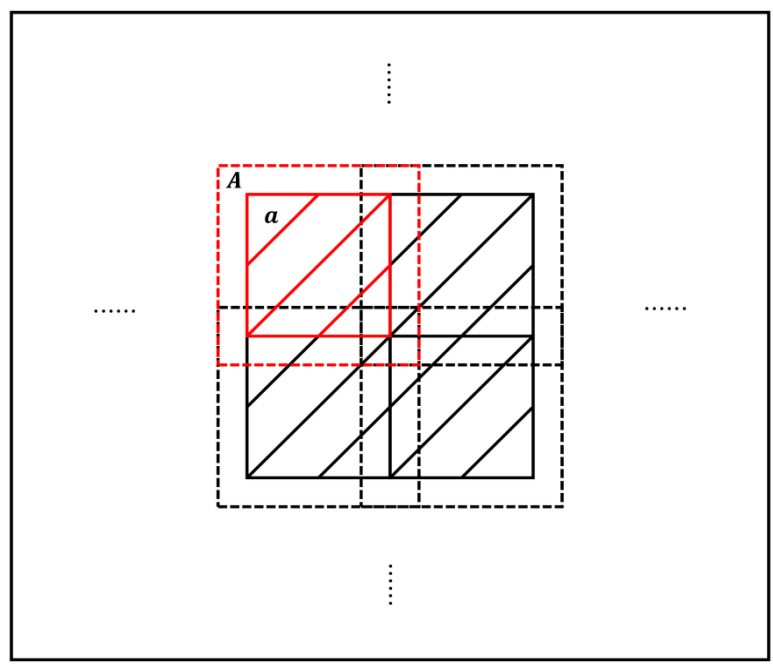

生成的小尺度图片更加清晰，并能够由网络快速计算出结果，最终得到整个遥感影像的识别结果。

点云文件生成的图片往往容易过亮或过暗，这是因为不同设备、不同路面材料造成的采集数据差异。往往会造成路面文字不清晰、标记不明显的问题。因此我们使用了直方图滤波算法，通过调整合适的参数，能够滤除强度小的点，保留强度大的点，转换为图像时通过高斯滤波，即可得到非常清晰的文字图像。

### 5.3.2 文字识别网络

传统 OCR 一般分为检测与识别两个阶段，由此，我们针对不同的环境、不同的语言进行调整，可以大大优化模型的性能。在经过上文滑窗边缘检测后，有文字的滑窗图片将被导入 paddleOCR 中进行含字符图片的截取和切割。

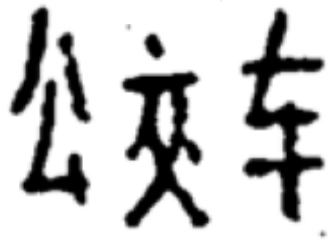

图 **5.1** paddleOCR 从点云图片中切割出的文字框

根据识别场景，可大致将 OCR 的识别阶段分为识别特定场景的专用 OCR 和识别多种场景的通用 OCR。通用 OCR，上文中的 paddleOCR 的识别部分可以用于更多的场景，也具有更大的应用范围。但由点云转化出的含文字图像布局多样、边缘失真严重、图片噪点众多，其基于手写体与印刷体数据集进行训练的模型识别效果很差。因此，我们决定基于 densenet 等成熟模型自行搭建二阶段识别模型。

鉴于点云文件中文字样本较少，我们结合点云转化图片中文字的特点，设计了该特定场景下数据集生成程序。字符范围选择了《现代汉语常用字表》中收录的 2500 个常用字和 1000 个次常用字，以及道路中可能出现的字母和数字。

在数据集生成尝试和文字识别过程中，我们逐渐摸索出了效果较好的对于文字图像的生成方法。我们先提取轮廓并依概率（进行过数次更改和优化）使其不规则，再进行中值滤波平滑边缘，并在图片中添加噪点，然后降低图像分辨率以符合点云转化图片的平均大小。

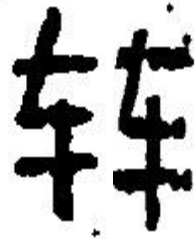

图 5.2 左侧为点云转化图片中二值化后的文字，右侧为生成的数据集

同时，因为点云中原有的文字太少，我们向部分点云图片中嵌入了部分文字，充实了验证集，并依此得到了不同生成方法识别的准确率。

表 5.2  不同数据集生成方法对应的识别准确率

| 边缘失真 | 噪点 | 是否中值滤波 | 识别准确率 |
| --- | --- | --- | --- |
| 无 | 无 | 无 | 5% |
| 有 | 无 | 无 | 55% |
| 有 | 有 | 1 次 | 75% |
| 优化后 | 有 | 1 次 | 85% |
| 优化后 | 有 | 2 次 | 90% |

# 第六章  结论与展望

综上所述，本文在路网构建、路标检测、多模态数据融合等方面提出了自己的见解，并试图解决一些数字孪生交通领域存在的问题。在一年的实践后，所有计划内的研究实验均完成，甚至在此基础上新增添了第五部分中的三维重建内容，预计今年九月至十一月期间发表两篇论文和一项专利。

放眼未来，数字孪生技术正在以惊人的速度推动以往碎片化数据管理流程的协作和连接，我们团队相信这项技术能够为人类揭开未来世界神秘面纱的一角，也能为人类发展带来更多福祉。

# 参考文献

[1]Lowe D G. Distinctive image features from scale-invariant keypoints[J]. International journal of computer vision, 2004, 60(2): 91-

[2]Sitzmann V., Zollhöfer, M., Wetzstein, G., 2019. Scene Representation Networks: Continuous 3D-Structure-Aware Neural Scene Representations. arXiv

[3]Tobin, J., Robotics, O., Abbeel, P., 2019. Geometry-Aware Neural Rendering. arXiv

[4]Hu H N, Cai Q Z, Wang D, et al. Joint monocular 3D vehicle detection and tracking[C]//Proceedings of the IEEE/CVF International Conference on Computer Vision. 2019: 5390-5399.

[5]Mattyus, G., Wang, S., Fidler, S., Urtasun, R., 2015. Enhancing road maps by parsing aerial images around the world, in: Proceedings of the IEEE International Conference on Computer Vision

[6]Ventura, C., Pont-Tuset, J., Caelles, S., Maninis, K.-K., Van Gool, L., 2018. Iterative Deep Learning for Road Topology Extraction. arXiv

[7]Paschalidou, D., Ulusoy, A.O., Schmitt, C., Gool, L. Van, Geiger, A., 2018. RayNet: Learning Volumetric 3D Reconstruction with Ray Potentials, in: 2018 IEEE/CVF Conference on Computer Vision and Pattern Recognition. IEEE, pp. 3897–3906.

[8]Homayounfar, N., Liang, J., Ma, W.C., Fan, J., Wu, X., Urtasun, R., 2019. DAGMapper: Learning to map by discovering lane topology, in: Proceedings of the IEEE International Conference on Computer Vision, IEEE. pp. 2911–2920

[9]Ye M, Xu S, Cao T. HVNet: Hybrid Voxel Network for LiDAR Based 3D Object Detection[J]. arXiv preprint arXiv:2003.00186, 2020

[10]Feng M, Hu S, Ang M H, et al. 2D3D-MatchNet: learning to match keypoints across 2D image and 3D point cloud[C]//2019 International Conference on Robotics and Automation (ICRA). IEEE, 2019: 4790-4796

[11]Pham Q H, Nguyen T, Hua B S, et al. JSIS3D: joint semantic-instance segmentation of 3d point clouds with multi-task pointwise networks and multi-value conditional random fields[C]//Proceedings of

the IEEE Conference on Computer Vision and Pattern Recognition. 2019: 8827-8836

[12]Ladicky, L., Saurer, O., Jeong, S., Maninchedda, F., Pollefeys, M., 2017. From Point Clouds to Mesh Using Regression, in: 2017 IEEE International Conference on Computer Vision (ICCV). IEEE, pp. 3913–3922.

[13]Zhong Y. Intrinsic shape signatures: A shape descriptor for 3d object recognition[C]//2009 IEEE 12th International Conference on Computer Vision Workshops, ICCV Workshops. IEEE, 2009: 689-696

[14] 姚连璧，屈宜琪，吴杭彬. 基于高精地图的道路场景三维建模[J]. 测绘科学技术，2022, 10(1): 1-12.

[15]王越，何伟，周琳,等. 基于车载 LiDAR 数据的高精道路地图制作[J]. 地理空间信息，2022, 20(6):4.

[16] CHEN Long ,LIU Kunhua, ZHOU Baoding, et al  Key technologies of multi‐agent collaborative high definition map construction [J] Acta Geodaetica et Cartog raphica Sinica ,2021.

[17] HE Yong , LU Hao , WANG Chunxiang , et al . Generation of precise lane‐level maps based on multi‐sensors [J] . Jour‐ nal of Chang'an University （Natural Science Edition ）2015,35(s1):274-278

[18] LIU C , JIANG K , YANG D , et al . Design of a multi‐ layer lane‐level map for vehicle route planning  [C ] 2017 .

[19]SHIMADAH , YAM AGUCHI A ， TAKADA H ， et al . Implementation and evaluation of local dynamic map in safety driving systems [J] . Journal of T ransportation Technologies,2015,5(2):102

[20] LIU Jingnan , ZHAN Jiao , GUO Chi , et al .Data logic structure and key technologies on intelligent high‐precision map [J] . Act Geodaetica et Cartographica Sinica , 2019,48(8):939-953

[21]刘通，高思洁，聂为之. 基于多模态信息融合的多目标检测算法[J]. 激光与光电子学进展，2022, 59(8):10.

[22] Q. Xu, X. Guan, J. Cao, Y. Ma and H. Wu, "MPR-GAN: A Novel Neural Rendering Framework for MLS Point Cloud With Deep Generative

Learning," in IEEE Transactions on Geoscience and Remote Sensing, vol. 60, pp. 1-16, 2022, Art no. 5704916, doi: 10.1109/TGRS.2022.3212389.

[23] Sarlin, P.E., DeTone, D., Malisiewicz, T. and Rabinovich, A., 2020. Superglue: Learning feature matching with graph neural networks. In Proceedings of the IEEE/CVF conference on computer vision and pattern recognition (pp. 4938-4947).

[24] D. Zermas, I. Izzat and N. Papanikolopoulos, "Fast segmentation of 3D point clouds: A paradigm on LiDAR data for autonomous vehicle applications," 2017 IEEE International Conference on Robotics and Automation (ICRA), Singapore, 2017, pp. 5067-5073, doi: 10.1109/ICRA.2017.7989591.

[25] Xu, G.; Pang, Y.; Bai, Z.; Wang, Y.; Lu, Z. A Fast Point Clouds Registration Algorithm for Laser Scanners. Appl. Sci. 2021, 11, 3426. https://doi.org/10.3390/app11083426

[26] M. Himmelsbach, F. v. Hundelshausen and H. .-J. Wuensche, "Fast segmentation of 3D point clouds for ground vehicles," 2010 IEEE Intelligent Vehicles Symposium, La Jolla, CA, USA, 2010, pp. 560-565, doi: 10.1109/IVS.2010.5548059.

[27] James Fossdyke, UK authorities plan 'digital roads' to make motorways 'safer and greener' (motor1.com)

[28] REDMON J, FARHADI A. YOLO9000: better, faster, stronger[C]//Proceedings of the IEEE conference on computer vision and pattern recognition, Honolulu, 2017: 7263-7271.

[29] REDMON J, FARHADI A. Yolov3: An incremental improvement[J].arXiv preprint arXiv:1804.02767, 2018.

[30] BOCHKOVSKIY A, WANG C Y, LIAO H Y M. Yolov4: Optimal speed and accuracy of object detection[J]. arXiv preprint arXiv:2004.10934, 2020.

[31]Y. Cao, J. Xu, S. Lin, F. Wei, and H. Hu, "GCNet: Non-local networks meet squeeze-excitation networks and beyond," ArXiv, vol. abs/1904.11492, 2019.

参考文献格式严格按照《武汉大学本科生毕业论文（设计）工作管理办法（修订）》标注